\documentclass[journal,9pt]{IEEEtran}
\usepackage{amsmath,amsfonts}
\usepackage{algorithmic}
\usepackage{algorithm}
\usepackage{array}
\usepackage[caption=false,font=normalsize,labelfont=sf,textfont=sf]{subfig}
\usepackage{textcomp}
\usepackage{stfloats}
\usepackage{url}
\usepackage{verbatim}
\usepackage{graphicx}
\usepackage{cite}
\usepackage{bm}
\usepackage{comment}
\usepackage{titlesec}

\begin{document}

\title{SmartQuant: CXL-based AI Model Store in Support of Runtime Configurable Weight Quantization}



\author{Rui Xie$^\dagger$, Asad Ul Haq$^\dagger$, Linsen Ma$^\dagger$, Krystal Sun$^\ddagger$,  Sanchari Sen$^*$, Swagath Venkataramani$^*$, Liu Liu$^\dagger$, and Tong Zhang$^\dagger$ \\$^\dagger$ Rensselaer Polytechnic Institute, NY, USA \hspace{6pt} $^\ddagger$ {\large BASIS Independent Fremont, CA, USA}, \hspace{6pt} $^*$ {\large IBM, NY, USA}

}



\maketitle
\begin{abstract}
Recent studies have revealed that, during the inference on generative AI models such as transformer, the importance of different weights exhibits substantial context-dependent variations. This naturally manifests a promising potential of adaptively configuring weight quantization to improve the generative AI inference efficiency. Although configurable weight quantization can readily leverage the hardware support of variable-precision arithmetics in modern GPU and AI accelerators, little prior research has studied how one could exploit variable weight quantization to proportionally improve the AI model memory access speed and energy efficiency. Motivated by the rapidly maturing CXL ecosystem, this work develops a CXL-based design solution to fill this gap. The key is to allow CXL memory controllers play an active role in supporting and exploiting runtime configurable weight quantization. Using transformer as a representative generative AI model, we carried out experiments that well demonstrate the effectiveness of the proposed design solution.
\end{abstract}


\section{Introduction}
With the wide acceptance of the scaling law~\cite{kaplan2020scaling,vaswani2017attention}, AI model size will continue to increase at least in the foreseeable future, despite today's AI models already contain over a trillion parameters~\cite{touvron2023llama, zhang2022opt}. To unleash the full potential of large AI models in the real world, it is essential to maximize the inference cost/energy efficiency, for which quantization is one of the most effective means and has been widely studied. To materialize the benefits of quantization, modern GPU and AI accelerators provide strong built-in hardware support for variable-precision arithmetics. For modern generative AI models such as transformer~\cite{vaswani2017attention}, recent studies show that the importance of different weights exhibits substantial context-dependent variations during inference~\cite{liu2023deja}. This observation opens new opportunities to improve the inference efficiency via adaptive weight quantization configuration. Although runtime weight quantization configuration can leverage the hardware support for variable-precision arithmetics in GPU and AI accelerators, little prior research has studied how one could exploit configurable weight quantization to  adaptively reduce the AI model load latency and energy consumption.

To fill this missing link, this letter presents a CXL-based AI model store that can gracefully support runtime configurable weight quantization. Built upon the rapidly maturing Compute Express Link (CXL) ecosystem with native support of memory pooling/sharing~\cite{sharma2023introduction}, CXL memory devices are well positioned to play a key role in future computing infrastructure. Since today's CPUs already have a large number of PCIe lanes~(e.g., AMD Genoa has 128 PCIe Gen5 lanes), it is reasonable to expect future CPU/GPU and AI accelerators could readily support 150+ PCIe lanes. Upcoming PCIe Gen7 specifies a per-lane bandwidth of 16GB/s, leading to total 4TB/s bandwidth~(2TB/s in each direction) with 128 PCIe lanes. In comparison, the upcoming 4th generation HBM~(high bandwidth memory) is expected to enable $\sim$1.5TB/s throughput per DRAM stack, and one GPU package may integrate 4$\sim$8 HBM DRAM stacks. Hence, CXL memory devices could be a viable cost-effective supplement~(or even alternative) to HBM for building future AI systems. This letter advocates for a CXL-based AI model store that can serve generative AI inference with dynamically configurable weight quantization~(e.g., FP16, FP8, FP6). Its practical implementation is subject to two major issues: (1)~How to materialize the potential of configurable weight quantization on improving the AI model DRAM access energy efficiency and throughput? As the straightforward design option, CXL memory controller reads the full-precision weights from DRAM and truncates/converts them into lower-precision ones, which however will not reduce the model load latency and energy consumption at all. (2)~How to simplify the system integration? Inference computing devices should be able to conveniently configure and load AI model weights from CXL memory devices. To address these issues, this letter presents a design solution called {\it SmartQuant} that consists of two simple yet effective methods, including
{\it bit-plane in-memory placement} and {\it memory logical space bloating}. Using the open-source transformer model~\cite{zhang2022opt} as a test vehicle and DRAM simulator DRAMSim3~\cite{li2020dramsim3}, we carried out experiments to study the effectiveness of the proposed design solution. In comparison to the straightforward design, the results show that, SmartQuant can reduce the model load latency by up to 42.1\% and memory access energy consumption by 40.3\%.

\section{Background and Motivation}
Compute Express Link (CXL) is a high-speed interconnect standard that enhances communication between computing engines and memory. Built on PCIe, CXL adds protocols for low-latency, high-bandwidth data transfer, crucial for AI and data-intensive tasks. It includes \textsf{CXL.io} for I/O, \textsf{CXL.cache} for caching in accelerators, and \textsf{CXL.mem} for flexible memory access. CXL's customizability and memory pooling support make it ideal for meeting the dynamic demands of modern computing~\cite{sharma2023introduction}.

Recognizing the paramount importance of quantization in AI inference (and even training), the chip design industry has made GPU and AI accelerators natively support variable-precision arithmetics so that their computational throughput and/or energy efficiency could gracefully scale with the data quantization precision. For example, latest Nvidia GPU Blackwell supports FP64/32/16/8/6/4 and INT8, and its peak performance is 2.25~PFLOPS~(peta FLOPS), 4.5~PFLOPS, and 9~PFLOPS under FP16, FP8, and FP4 quantization, respectively. This capability lays a solid foundation for algorithms/software to navigate the trade-off space between the AI inference/training quality and implementation cost by dynamically adjusting the data quantization.

As a topic of great current interest, generative AI models such as transformer~\cite{vaswani2017attention} have been most successfully demonstrating the power of the scaling law~\cite{kaplan2020scaling}. Leading-edge generative AI models now boast over a trillion parameters, with no indication of slowing down in the continuous growth of model size. This has led to growing attention on improving their training/inference efficiency. Motivated by the observation that the importance of different weights exhibit significant context-dependent variations during inference on generative AI models, recent work proposed to reduce the inference cost by selectively bypassing less important model weights~\cite{liu2023deja}. This can be naturally extended to more flexible and fine-grained configuration of weight quantization in adaptation to their context-dependent importance.  This will enable generative AI inference more gracefully explore the quality vs.~cost trade-off space. Although it could seamlessly leverage the existing variable-precision arithmetic support of modern GPU and AI accelerators, little prior work has studied how one could exploit weight quantization configuration to improve the model memory load speed and access  efficiency. 

\section{Proposed Design}
This section presents a solution that enables future CXL memory devices to effectively support context-dependent fine-grained weight quantization configuration for generative AI inference. First, we note that, to facilitate its practical implementation, weight quantization should be configured with a coarse granularity: A large chunk of closely related model weights~(e.g., all the weights associated with the same attention head in a transformer model) are always assigned with the same runtime contextual importance and hence the same quantization precision. As a result, runtime model weight quantization configuration only varies from one large chunk to another. Therefore, inference computing devices fetch model weights from DRAM chunk-by-chunk, where all the weights within each chunk always have the same quantization configuration. Note that different chunks may contain different number of weights. Such {\it chunked} quantization configuration not only simplifies the runtime evaluation of contextual weight importance, but also streamlines the process of loading model from memory and facilitates the utilization of variable-precision arithmetic hardware support in modern GPU and AI accelerators.

\begin{figure}[htbp]
	\centering
    \includegraphics[width=0.9\linewidth]{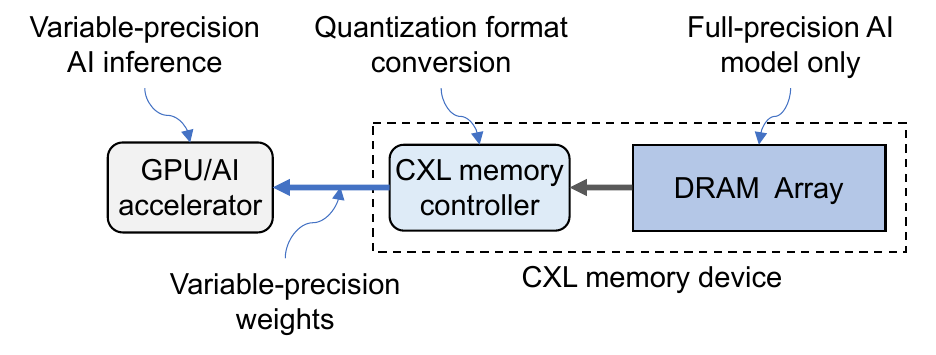}
	\caption{Illustration of CXL memory devices with built-in quantization format conversion to support low-cost AI inference.}
\label{fig.overview}
\end{figure}

Let $\mathcal{Q}$ denote the set of supported quantization formats,~e.g., $\mathcal{Q}=$\{FP16, FP8, FP6, FP4\}. The most convenient option is to store multiple copies of the AI model in DRAM, each copy corresponding to one quantization format $q_l\in \mathcal{Q}$, similar to the storage of different-quality video files in support of adaptive bit rate streaming~\cite{bentaleb2018survey}. This however incurs a significant memory capacity usage overhead because of the very large generative AI model size. Intuitively, one alternative is to store only the full-precision AI model in memory and construct reduced-precision model weights via on-the-fly quantization format conversion. The controller chip inside each CXL memory device can naturally serve this format-conversion task, as illustrated in Fig.~\ref{fig.overview}. When fetching each chunk of model weights, inference computing devices specifies the intended quantization format $q_l\in \mathcal{Q}$, and CXL memory devices internally on-the-fly construct the $q_l$-quantized weights from the stored full-precision version. When implementing such CXL-based AI model store with built-in quantization format conversion capability, we should aim at achieving two goals: 
\begin{enumerate}
\item {\it Proportional DRAM access efficiency}: Let $\alpha>1$ denote the average weight precision reduction ratio~(i.e., the ratio between the total number of bits in the full-precision model and dynamically quantized model). Ideally, the DRAM access efficiency should improve by $\alpha\times$, i.e., both the AI model load energy consumption and latency reduces by $\alpha\times$.
\item {\it Seamless system integration}: The host should be able to conveniently convey the intended quantization format information to CXL memory devices while being 100\%-compliant to the \textsf{CXL.mem} protocol. To ensure the practical feasibility of the proposed design solution, we should restrain from adding any new command \textsf{CXL.mem} protocol commands. 
\end{enumerate}
To accomplish these two goals, we develop two simple yet effective techniques that will be described in the remainder of this section.

\subsection{Bit-Plane In-Memory Placement} \label{subsection.bit-plane in-memory placement}

In conventional practice, all the bits of each numeric value are always stored contiguously together in memory. This however will prevent CXL memory devices from achieving proportional DRAM access efficiency in support of configurable weight quantization: Assume we store each full-precision weight as a whole in DRAM. Due to the per-page DRAM cell activation and DRAM access burst length, regardless of the target precision, CXL memory controller must always fetch full-precision weights from DRAM to realize the quantization format conversion. As a result, CXL memory devices always have a fixed quantization-independent, other than the desired quantization-proportional, DRAM access efficiency. 

\begin{figure}[htbp]
	\centering
    \includegraphics[width=.9\linewidth]{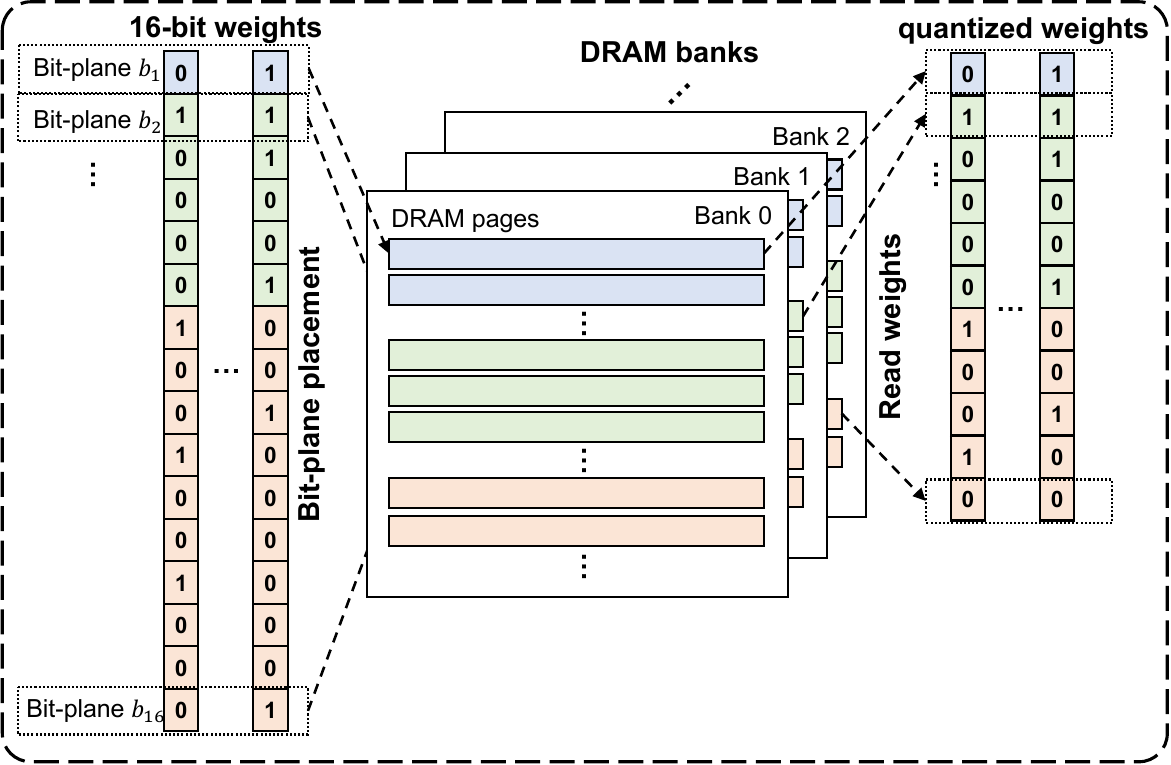}
	\caption{Illustration of DRAM structure with bit-plane in-memory placement.}
\label{fig.bitplane illustration}
\end{figure}

To address this issue, the only viable option is to bit-wise disaggregate the in-memory storage of each full-precision weight, similar to a bit-wise column-store. Let $N_1$ denote the number of bits in full-precision weight quantization. We store all the model weights in $N_1$ bit-planes, each bit-plane corresponding to one bit position in the $N_1$-bit full-precision quantization. All the $N_1$ bit-planes are stored in DRAM independently from each other, as illustrated in  Fig.~\ref{fig.bitplane illustration}. Through such bit-plane in-memory placement, CXL memory controller can selectively fetch a {\it just-enough} portion of each \textit{chunked} full-precision weights from DRAM to realize quantization format conversion. Suppose the full-precision quantization contains 1-bit sign, $n_{e}$-bit exponent, and $n_{m}$-bit mantissa~(i.e., $N_1=1+n_e+n_m$). Assume inference computing devices request a large chunk of weights under a reduced-precision quantization with 1-bit sign, $r_{e}$-bit exponent, and $r_{m}$-bit mantissa. We could perform quantization format conversion by fetching data from $1+(r_e+d_e)+(r_m+d_m)$ in-memory bit-planes, where $d_e, d_m\ge 0$. When setting $d_e=d_m=0$, we obtain the reduced-precision model weights by directly truncating the full-precision model weights. When setting $d_e$ and/or $d_m$ as 1 or 2, we can improve the quantization format conversion accuracy by rounding-off the full-precision model weights. Utilizing such bit-plane in-memory weight placement, we could proportionally reduce the amount of DRAM page activation and data transfer, leading to the proportional DRAM access efficiency.

\vspace*{-12pt}
\subsection{Memory Logical Space Bloating}
\begin{figure}[htbp]
	\centering
    \includegraphics[width=.9\linewidth]{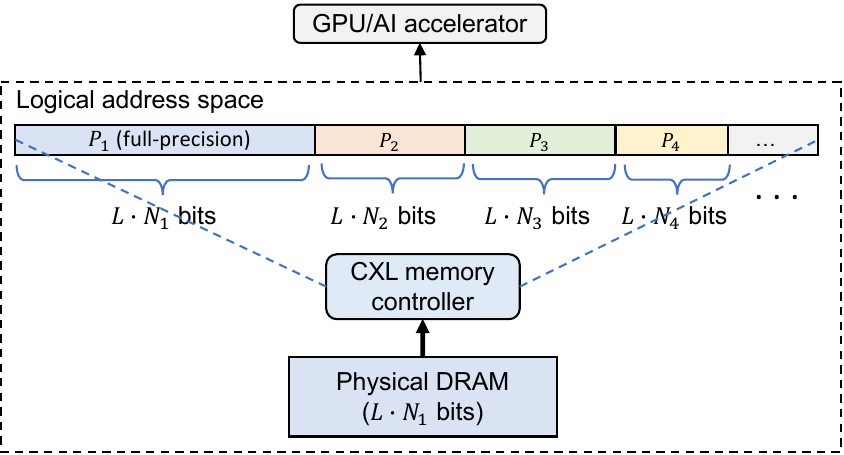}
	\caption{Illustration of CXL memory logical space bloating.}
\label{fig.memory virtualization}
\end{figure}

To facilitate host inference computing devices fetch model weight chunks with configurable quantization, CXL memory devices could expose a {\it bloated} logical memory space: Let $L$ denote the total number of model weights, $s$ denote the number of different quantization formats, and $N_i$ denote the number of bits in the $i$-th quantization format~($N_1$ corresponds to the full-precision quantization). CXL memory devices exposes a logical space consisting of $s$ regions, where each region $P_i$ corresponds to the $i$-th quantization format and has the capacity of $L\cdot N_i$ bits. As illustrated in Fig.~\ref{fig.memory virtualization}, each region $P_i$ spans over a continuous logical memory space of $L\cdot N_i$ bits, and all the $s$ regions altogether span over a continuous logical memory space of $\sum_{i=1}^s L\cdot N_i$ bits. In contrast, CXL memory devices internally only have ($L\cdot N_1$)-bit physical DRAM that stores the full-precision AI model. This can be further illustrated in Fig.~\ref{fig.memory virtualization}. 

Upon the bloated logical memory space, host inference computing devices can conveniently issue requests to fetch model weight chunks with different quantization formats. To fetch a length-$l$ chunk of model weights with the $i$-th quantization format, host inference computing devices simply issues a size-($l\cdot N_i$) sequential read request to the memory region $P_i$. Upon receiving the memory read request, CXL memory controller internally fetch the corresponding data from $N_i$ (or few more) bit-planes and accordingly construct the requested length-$l$ chunk of model weights with the $i$-th quantization format. 

\vspace*{-12pt}
\section{Evaluation}

\subsection{Runtime Configurable Weight Quantization}
Using the open-source transformer model OPT~\cite{zhang2022opt}, we first studied the effect of configurable weight quantization on the inference quality. All the experiments were done on a server with 8 Nvidia L40 48GB GPUs. Following the recent work~\cite{liu2023deja}, we use a small-size offline-trained neural network as a {\it predictor} on each layer to estimate the relative importance of each attention head and neuron. Taking the activation from the previous layer as input, each predictor produces an importance score in the range of $[0, 1]$ 
for each attention head and 
each neuron, where 0 and 1 correspond to the least and highest importance.
Given the total $s$ quantization formats, the system can adjust a set of $s-1$ thresholds $\{t_1, t_2, \ldots, t_{s-1}\}$ to realize the runtime configurable weight quantization.

\begin{figure}[htbp]
	\centering
    \includegraphics[width=\linewidth]{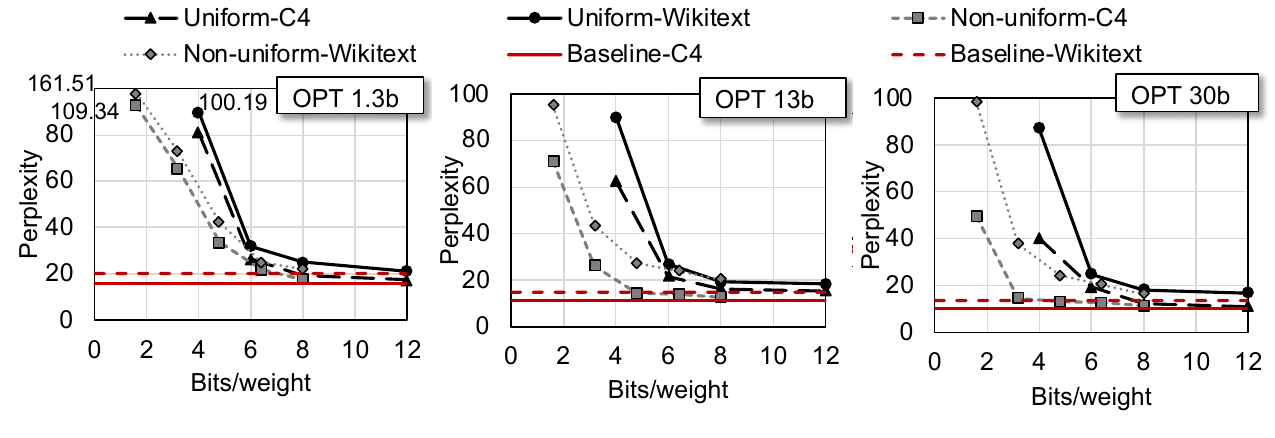}
	\caption{Measured perplexity~(lower is better) under  different weight quantization configurations.}
\label{fig.perplexity}
\end{figure}

We considered three different scenarios: (i)~{\it Baseline}: All the OPT model weights use the full-precision quantization of FP16; (ii)~{\it Uniform}: All the OPT weights use the same quantization precision; (iii)~{\it Non-uniform}: Given the target average number of bits per weight, the system dynamically configure weight quantization based on the estimated importance score. 
In this work, we set $s=6$ with the quantization formats of FP16, FP12, FP8, FP6, FP4, and FP0~(i.e., skipped weight). We use three different OPT models including OPT 1.3b, OPT 13b, and OPT 30b, and evaluate the inference perplexity using the language modeling datasets C4~\cite{raffel2020exploring} and WikiText~\cite{merity2016pointer}. 
Fig.~\ref{fig.perplexity} shows the measured perplexity under different average number of bits per weight (note that the {\it baseline} always use the full-precision FP16 on all the weights). The results show that, under the same average number of bits per weight, non-uniform dynamic quantization consistently achieves lower~(i.e., better) perplexities than uniform quantization. This can be intuitively justified by the significant context-dependent weight importance variation observed in prior work~\cite{liu2023deja}. 
Moreover, the results show that non-uniform dynamic quantization can be more effective as AI models become bigger. For example, under the C4 dataset, the perplexity gap between {\it non-uniform} with average of 8bits/weight and {\it baseline} is 13.0\% (i.e., 17.4 vs.~15.4), while the gap reduces to 11.6\% in the OPT 30b model~(i.e., 11.5 vs.~10.3).

\vspace*{-12pt}
\subsection{DRAM Access Efficiency}
\begin{figure}[htbp]
	\centering
    \includegraphics[width=\linewidth]{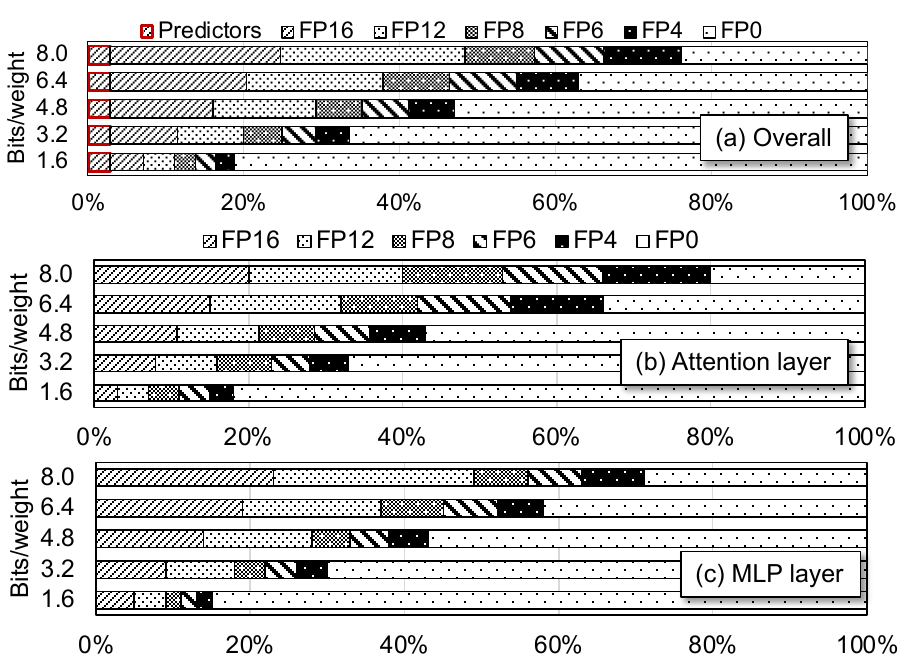}
	\caption{Average percentage of different quantization formats in (a) the entire model (including all the predictors), (b) attention layers, and (c) MLP layers under different target bits/weight in OPT 30b model.}
\label{fig.percentage}
\end{figure}

Using the open-source DRAM simulator DRAMSim3~\cite{li2020dramsim3}, we further evaluated the DRAM access efficiency of configurable weight quantization. We set that each CXL memory module contains 4 DRAM channels, each channel hosting 10 $\times$4 DDR5-4800 devices. We collected the memory access traces during the inference of OPT 30b model on the WikiText dataset, which was fed into the DRAM simulator. We studied two different scenarios: (i) \textit{Traditional}: As the straightforward design option, the full-precision AI model are stored in DRAM weight-by-weight. To support dynamic weight quantization, the CXL memory controller reads full-precision weights from DRAM and then converts them into lower-precision formats; (ii) \textit{SmartQuant}: As described in Sec.~\ref{subsection.bit-plane in-memory placement}, this approach utilizes bit-plane placement to minimize quantization overhead, enabling the CXL memory controller to read only the required bits from the full-precision weights in DRAM. All the predictors' weights are always fetched in the full-precision FP16 format.
Fig.~\ref{fig.percentage}(a) shows six different quantization formats and predictor distribution in entire OPT 30b model. As the average number of bits per weight increases from 1.6 to 8.0, the proportion of predictors decreases from 15.2\% to 3.7\%. 
Fig~\ref{fig.percentage}(b) and (c) shows the average percentages of the quantization formats across attention and MLP layers under different target average number of bits per weight. 


\subsubsection{DRAM Access Energy}

\begin{figure}[htbp]
\centering\includegraphics[width=\linewidth]{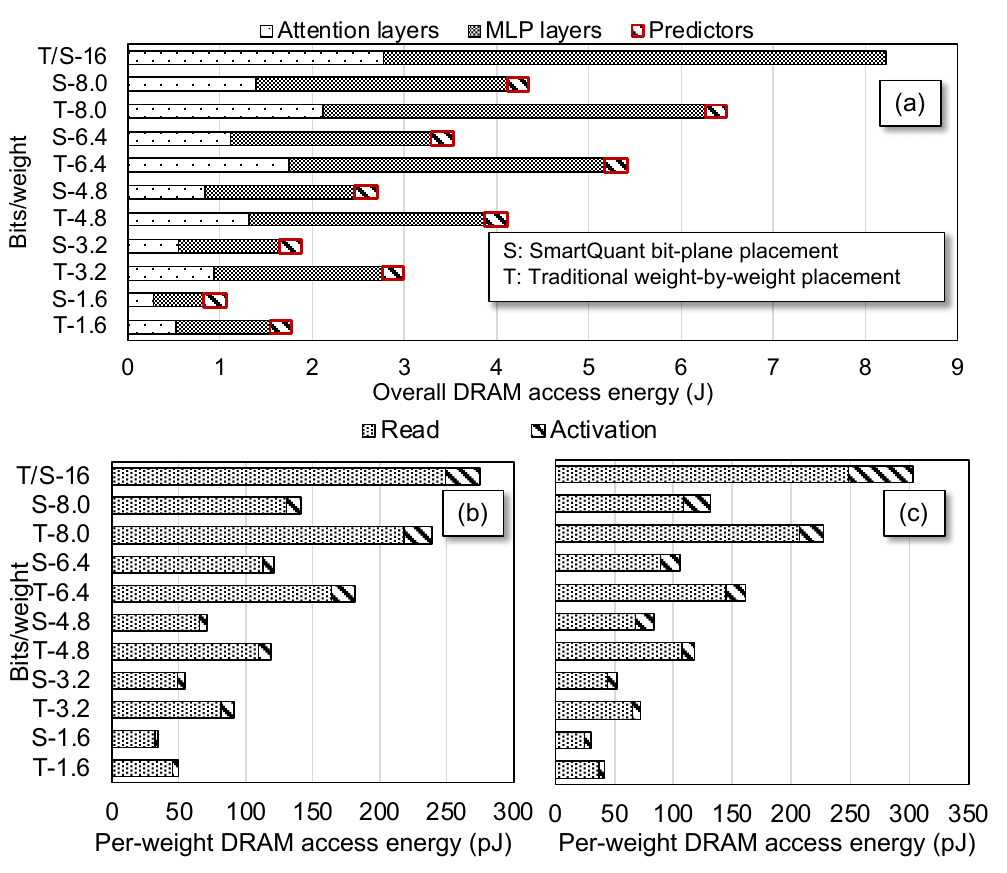}
	\caption{Comparison of (a) total DRAM access energy consumption when loading the entire model once, and per-weight DRAM access energy consumption of (b) attention heads and (c) MLP layers under different target bits/weight in OPT 30b model.}
\label{fig.DRAM access energy}
\end{figure}

Fig.~\ref{fig.DRAM access energy}(a) shows the total DRAM access energy when loading the entire AI model once, with the energy consumption breakdown among attention layers, MLP layers, and predictors. Compared to the \textit{traditional} approach, \textit{SmartQuant} reduces the DRAM access energy by up to 40.3\%. 
Fig.~\ref{fig.DRAM access energy}(b) and (c) shows the average per-weight DRAM access energy consumption for reading weights in an attention head and an MLP neuron of the OPT 30b model. We treat all weights associated with the same attention head or MLP neuron as a single chunk and hence assign the same quantization format. In the OPT 30b model, this corresponds to $3.7\times10^6$ weights in an attention head and $7.2\times10^3$ weights in an MLP neuron. Both the {\it traditional} and {\it SmartQuant} approaches directly skip all the chunks with FP0 and fetch the same amount of data from DRAM for chunks with the full-precision FP16 quantization. For chunks with reduced-precision quantization formats, {\it SmartQuant} fetches less amount of data from DRAM than the {\it traditional} approach, leading to energy and latency reduction. Fig.~\ref{fig.DRAM access energy} clearly shows the DRAM access energy saving of {\it SmartQuant}. For attention heads, under target bits/weight of 1.6, 4.8, and 8.0, the per-weight DRAM access energy is 49.6pJ, 118.9pJ, and 238.9pJ when using the {\it traditional} approach, which reduce to  34.5pJ, 70.8pJ, and 141.2pJ under {\it SmartQuant}, representing 30.5\%, 40.4\%, and 40.9\% reduction, respectively. Similarly, for MLP layers, under target bits/weight of 1.6, 4.8, and 8.0, compared with the {\it traditional} approach, {\it SmartQuant} could reduce the per-weight DRAM access energy by 19.4\%, 20.3\%, and 33.9\%, respectively. 

\vspace*{-12pt}

\vspace*{-5pt}
\subsubsection{Model Load Latency}
\begin{figure}[htbp]
	\centering
    \includegraphics[width=\linewidth]{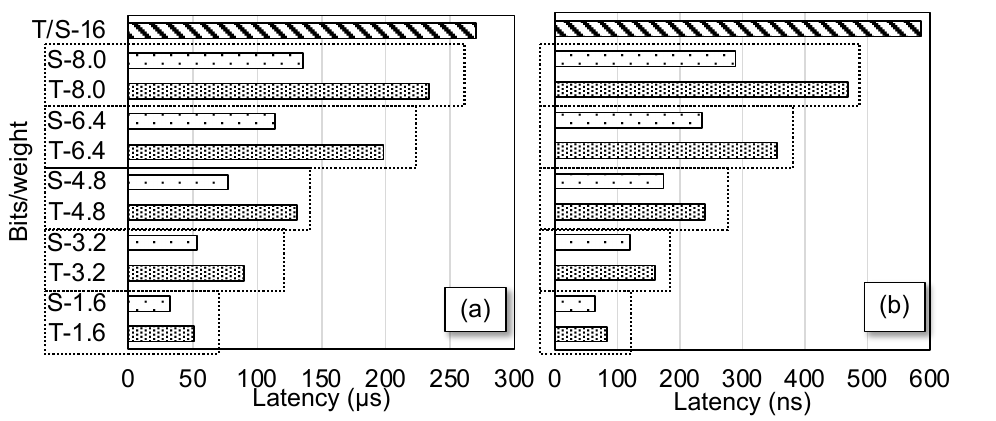}
	\caption{Average (a) attention head and (b) MLP neuron normalized load latency across different non-uniform quantized settings on WikiText using OPT 30b.}
\label{fig.read latency}
\end{figure}

We further measured the AI model load latency when using different design approaches. Fig.~\ref{fig.read latency} shows the average load latency of each attention head and each neuron in OPT 30b model. The results clearly show the effectiveness of the bit-lane in-memory placement on reducing the model load latency. For each attention head that contains $3.7\times10^6$ weights, under target bits/weight of 1.6, 4.8, and 8.0, the load latency is 50.7$\mu$s, 89.9$\mu$s, and 233.9$\mu$s when using the {\it traditional} approach, which reduce to 32.4$\mu$s, 77.4$\mu$s, and 135.6$\mu$s when using {\it SmartQuant}, representing 36.2\%, 40.6\%, and 42.1\% reduction, respectively. Similarly, for each neuron in MLP layers, under target bits/weight of 1.6, 4.8, and 8.0, compared with the {\it traditional} approach, {\it SmartQuant} could reduce the load latency by 24.8\%, 27.9\%, and 38.4\%, respectively.  

\vspace*{-5pt}
\section{Conclusion}
This letter introduces {\it SmartQuant}, a CXL-based AI model storage solution that can exploit runtime context-dependent model weight importance variation to proportionally reduce model access energy consumption and load latency. The CXL memory controller is responsible for on-the-fly weight quantization conversion to supply variable-precision model weights. It employs a weight bit-plane in-memory placement strategy to enable fetching {\it just-enough} bits from DRAM in adaptation to desired weight quantization precision. Furthermore, it simplifies the system integration by exposing a bloated logical memory space that is partitioned into regions with different quantization precisions. Using open-source transformer model, we carried out experiments to demonstrate the effectiveness of context-dependent model weight quantization configuration and the proposed design solution on reducing the model DRAM access energy consumption and load latency. 


%
\vspace*{-10pt}
\bibliographystyle{IEEEtran}
\bibliography{ref}

\end{document}